\documentclass[letterpaper]{article} % DO NOT CHANGE THIS
\usepackage{aaai2027}
\usepackage[hyphens]{url}  % DO NOT CHANGE THIS
\usepackage{graphicx} % DO NOT CHANGE THIS
\usepackage{natbib}  % DO NOT CHANGE THIS AND DO NOT ADD ANY OPTIONS TO IT
\usepackage{caption} % DO NOT CHANGE THIS AND DO NOT ADD ANY OPTIONS TO IT
\usepackage{algorithm}
\usepackage{algorithmic}
\usepackage{amsmath}
\usepackage{makecell}
\usepackage{amssymb}
\usepackage{multirow}
\usepackage[table]{xcolor}
\usepackage{comment}
\usepackage{wasysym}
\usepackage[switch]{lineno}
\usepackage{cleveref} % 通常放在 hyperref 之后
\usepackage{newfloat}
\usepackage{listings}
\DeclareCaptionStyle{ruled}{labelfont=normalfont,labelsep=colon,strut=off} % DO NOT CHANGE THIS
\floatstyle{ruled}
\newfloat{listing}{tb}{lst}{}
\floatname{listing}{Listing}

\usepackage{booktabs}

\title{MSGAT: Multi-Head Spiking Graph Attention  with Similarity-Space Fusion for Image-Text Retrieval}
\author{
    Xintao Zong\textsuperscript{1}, \quad
    Wenxuan Liu\textsuperscript{1}, \quad
    Jianhao Ding\textsuperscript{1}, \quad
    Zhaofei Yu\textsuperscript{1}\corresponding, \quad
    Tiejun Huang\textsuperscript{1}\corresponding \\
    \textsuperscript{1}Peking University \\
}
\affiliations{
}

\begin{document}
%\linenumbers
\maketitle

\begin{abstract}

Spiking neural networks (SNNs) offer an energy-efficient computing paradigm through sparse event-driven computation, showing great potential for efficient multimodal learning. However, applying SNNs to high-level multimodal tasks, such as image-text retrieval (ITR), remains challenging, since sparse spike representations make it difficult to capture semantic structures required for cross-modal alignment. Existing spiking ITR methods rely on local alignment and additional soft-label supervision during training, while lacking awareness of structural and multi-granularity relationships. To address these issues, we propose a Multi-head Spiking Graph Attention Network (\textbf{MSGAT}) for structural modeling and equip it with dynamic attention heads to capture complementary relational patterns and enable spike-driven graph reasoning and aggregation. However, within a two-branch multi-granularity fusion framework, the fine-grained spike representations generated by MSGAT are sparse and discrete, whereas the global representations are continuous, making conventional feature-level fusion susceptible to interference across heterogeneous representations. Therefore, we introduce \textbf{Sim-Fuse}, a similarity-space fusion alignment strategy integrating coarse- and fine-grained matching relations while avoiding direct fusion of heterogeneous representations. Experiments on Flickr30K and MSCOCO show our method outperforms ANN methods under matched settings and existing SNN retrieval baselines. Moreover, with only two time steps, our SNN achieves comparable or superior performance to its ANN counterpart while reducing theoretical module-level energy by 55\%. The code is provided in the Supplementary Materials.
\end{abstract}

\section{Introduction}

Image-text retrieval (ITR) aims to identify semantically corresponding samples across visual and textual modalities~\cite{VSE-infty}. Recent ANN-based methods have achieved remarkable performance by global semantic alignment~\cite{USER}, fine-grained region-word correspondence~\cite{CHAN}, multi-level matching strategies~\cite{HREM}, and large-scale pre-trained models~\cite{CUSA}. However, existing methods typically rely on deep architectures, dense floating-point operations, and complex cross-modal interactions~\cite{SCAN}, incurring considerable computational and energy costs that limit scalability  in large-scale retrieval scenarios.
%However, many fine-grained matching methods still depend on deep architectures, dense floating-point operations, and complex cross-modal interactions~\cite{SCAN}, incurring considerable computational and energy costs that limit scalability for large image and text collections in practical retrieval systems.
\begin{figure}[!t]
	\centering
	\includegraphics[width = \linewidth]{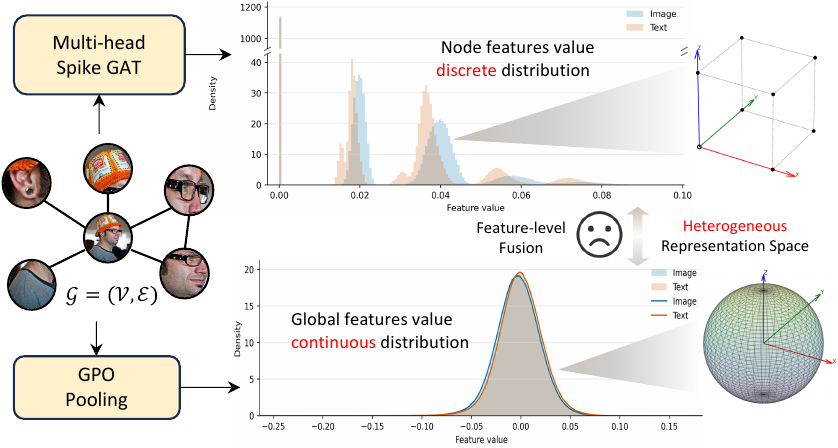} 
	\caption{Motivation. Sparse, discrete SNN representations and continuous representations produced by ANN-based GPO (Generalized Pooling Operator) lie in heterogeneous spaces, making conventional feature-level fusion ineffective.}
	\label{fig:motivation}
\end{figure}
Spiking neural networks (SNNs) provide a promising alternative for efficient multimodal learning through sparse spike activations and event-driven computation. Recently, CMSF~\cite{CMSF} established the feasibility of  SNNs for cross-modal retrieval. Nevertheless, CMSF primarily relies on local spike encoding and additional soft-label supervision during training, leaving semantic relation reasoning and global semantic alignment underexplored, although both are important for multimodal understanding and alignment.
%Recently, CMSF~\cite{CMSF} demonstrated the feasibility of directly trained SNNs for cross-modal retrieval for the first time. Nevertheless, CMSF primarily focuses on local spike encoding and soft-label guided optimization, while semantic relation reasoning and global  alignment remains unexplored, which is essential for multimodal understanding and alignment.

High-level cross-modal alignment requires modeling structural relationships beyond isolated visual regions and textual tokens. These semantic units are inherently connected through visual and linguistic dependencies, which provide important cues for distinguishing fine-grained semantic correspondence. \textit{Therefore, spiking ITR calls for an effective  spike-driven graph reasoning mechanism that can capture structural dependencies and enable semantic propagation}.

Fine-grained local features and coarse-grained global features provide complementary information for multi-granularity image-text alignment. Existing ANN-based ITR methods commonly integrate these representations through feature-level fusion. However, such a strategy introduces another challenge in SNNs, as illustrated in Fig.~\ref{fig:motivation}: fine-grained representations from SNNs exhibit sparse and discrete characteristics, while global representations from GPO~\cite{VSE-infty} remain continuous. \textit{Conventional feature-level fusion across these heterogeneous representation spaces may introduce interference and destabilize joint optimization.} %\textit{The inconsistency  between these heterogeneous representation spaces can cause representation interference and destabilize joint optimization.}

To address these challenges, we propose a Multi-head Spiking Graph Attention Network (\textbf{MSGAT}) with Similarity Space Fusion (\textbf{Sim-Fuse}) alignment strategy for ITR. Specifically, MSGAT constructs spiking visual and textual graphs and performs spike-driven graph reasoning and  aggregation, enabling fine-grained structural semantic modeling. By employing multiple dynamic attention heads, MSGAT captures complementary relational patterns from different representation subspaces while preserving the event-driven nature of SNNs. Furthermore, we introduce Sim-Fuse, which integrates coarse- and fine-grained matching relations at the similarity level, avoiding feature-level fusion between the discrete and continuous representations produced by MSGAT and GPO, respectively. This similarity-space fusion enables multi-granularity semantic integration. %Together, MSGAT and Sim-Fuse form an efficient spiking retrieval framework that enhances semantic alignment capability while maintaining the computational advantages of SNNs.
Together, MSGAT and Sim-Fuse form an efficient spiking retrieval framework that improves multi-granularity semantic alignment while retaining the computational advantages of SNNs.

%Experiments on Flickr30K and MSCOCO datasets demonstrate that our method outperforms state-of-the-art ANN-based methods and previous SNN retrieval baselines. Moreover, with only 2 time steps, our SNN achieves comparable or even superior performance to its ANN counterpart while reducing energy consumption by 55\%.
Under comparable feature-extraction settings, experiments on Flickr30K and MSCOCO demonstrate that our method outperforms ANN-based methods and existing SNN retrieval baselines. Moreover, with only two time steps, our proposed MSGAT achieves performance comparable or even superior to that of its ANN counterpart while reducing the theoretical inference energy of the graph-reasoning module by approximately 55\%.

The main contributions are summarized as follows:
\begin{itemize}
    
   \item We propose MSGAT, a novel spike-driven multi-head graph attention network that introduces dynamic structure-aware reasoning into image-text retrieval. By exploiting graph relationships and complementary attention heads, MSGAT enables fine-grained semantic graph  propagation under sparse  event-driven computation.

\item We introduce Sim-Fuse, a similarity-space fusion alignment strategy for SNNs that integrates coarse- and fine-grained matching relations while bypassing heterogeneous feature representation spaces, effectively improving multi-granularity semantic alignment.

\item Extensive experiments on Flickr30K and MSCOCO demonstrate that the proposed framework outperforms state-of-the-art methods. With only two time steps, our model achieves competitive performance with ANN counterparts while reducing energy consumption by 55\%. 
\end{itemize}

\section{Related Works}
\subsection{Image-Text Retrieval}
Recent image-text retrieval methods have explored graph structures to model semantic relations beyond isolated regions or tokens. VSRN~\cite{VSRN} performs graph convolutional reasoning over bottom-up image regions to obtain relation-enhanced visual features. GSMN~\cite{GSMN} constructs structured visual and textual graphs whose nodes correspond to objects, relations, or attributes, and learns fine-grained phrase correspondence through node-level and structure-level matching. CSGM~\cite{CGMN} formulates retrieval as object-level and relationship-level matching between visual and textual scene graphs. HREM~\cite{HREM} further models fragment-level and instance-level semantic relations with a cross-embedding association graph to improve hard negative discrimination. CORA~\cite{CORA} represents captions as scene graphs and uses graph attention networks to encode object-attribute and object-object relations in a dual-encoder framework. CHAN~\cite{CHAN} emphasizes discriminative hard alignments to reduce noisy fine-grained matching. %Although these methods show the effectiveness of graph reasoning for image-text retrieval, they rely on ANN-based continuous  and floating-point computation. 
CMSF~\cite{CMSF} first applies SNNs to ITR, demonstrating their feasibility for semantically demanding  complex multimodal tasks. However, it relies on soft-label supervision and suffers from interference by semantically irrelevant nodes in self-attention. Our proposed MSGAT mitigates these limitations via spike-driven graph reasoning, graph aggregation, multi-head dynamic attention, and multi-granularity fusion.  %However, they are mainly developed for dense ANN features, where feature-level fusion and attention interaction are relatively stable. %In SNN-based retrieval, pooled spike embeddings can become over-smoothed due to sparse binary activations, making direct coarse- and fine-grained feature fusion unstable. Therefore, we fuse global and fine-grained matching relations in the similarity-matrix space, avoiding direct perturbation of spike feature embeddings while providing stable multi-granularity supervision.

\subsection{Spiking Neural Networks}
Early work such as~\cite{Cao_CNN2SNN} explored converting deep CNNs into SNNs by interpreting ANN activations as firing rates. \cite{LV-SpikeCNN-test} proposed a two-stage pipeline combining ANN-to-SNN conversion with fine-tuning. Parallel efforts investigated direct training with surrogate gradients \cite{wu-STBP} .  Spikformer~\cite{Zhou_spikformer} and its variants~\cite{QKformer,Spike-Driven-transformer-V2} proposed a spike-driven self-attention mechanism, marking a milestone in the development of SNNs. Recently, SNNs have achieved notable progress in unimodal visual tasks, such as image classification~\cite{ONESTEP}, object detection~\cite{SUYOLO,spikeyolo}, semantic segmentation~\cite{spike2former}, and saliency detection~\cite{liuwenxuan}.
%Spiking neural networks (SNNs) provide an energy-efficient computing paradigm by using sparse binary spikes for information transmission. Compared with conventional artificial neural networks based on dense floating-point activations, SNNs communicate through event-driven spike signals and are therefore suitable for resource-constrained scenarios. 
Some works have extended SNNs to graph-structured learning. SpikingGCN~\cite{SpikeGCN} encodes graph structures into spike sequences with graph convolution operation, GSAT~\cite{GSAT} learns spike-based graph attention for robust  representation learning, SpikeGCL~\cite{SpikeGCL} introduces graph contrastive learning into SNNs to obtain compact binary representations, and SpikeNet~\cite{SpikeNet} explores spiking neurons for dynamic  representation learning. SpikingHAN~\cite{SpikingHAN} further applies SNNs to heterogeneous graph learning with a shared convolution layer. Our  MSGAT is tailored to  multimodal ITR  and differs from  prior graph static-attention methods by introducing multi-head dynamic attention with softmax-free spike aggregation to capture structural semantic representations.%Our proposed MSGAT is tailored to the multimodal ITR task and introduces dynamic attention with softmax-free multi-head spike aggregation to capture fine-grained structural semantic representation.

%Different from existing spiking graph attention methods that still compute continuous attention scores before converting them into spike representations, our SGAT performs attention generation and neighborhood aggregation directly in the binary spike domain. Specifically, the interaction between spike queries and spike keys is implemented through 0-1 spike operations and graph masks, and the subsequent message passing is restricted by the binary adjacency structure. Therefore, the proposed SGAT avoids dense softmax attention and continuous attention coefficients in the graph reasoning stage, making it more consistent with sparse event-driven computation.

\section{Proposed Method}
\begin{figure*}[!t]
	\centering
	\includegraphics[width = \linewidth]{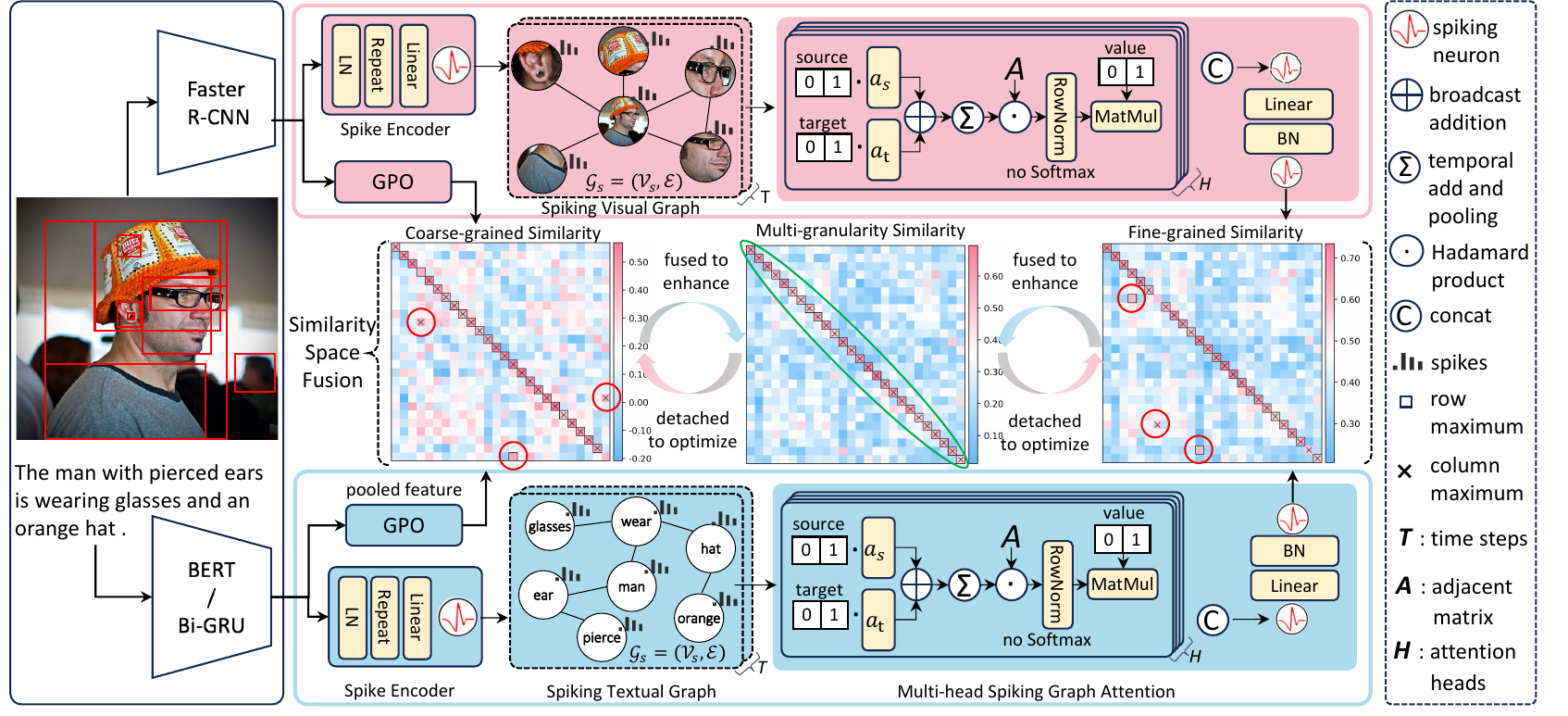} % 图像文件名
	\caption{Framework of  MSGAT for ITR. Left: Replaceable extractors produce floating-point node features. Top and bottom: GPO derives global features, Spike Encoder converts node features into spikes for Spiking Graph construction. MSGAT performs spike-driven graph reasoning and aggregation to obtain neighborhood-aware spiking fine-grained features. Middle: Coarse- and fine-grained similarities are fused into a detached multi-granularity target, which in turn aligns and optimizes the two branches. }
	\label{fig:framework}
\end{figure*}

%The overview of our proposed network is illustrated in~\cref{fig:framework}. We first extract floating-point node features of image and text, and then construct spiking visual and textual graph, which are processed by two complementary branches: MSGAT captures fine-grained structural semantics through spike-driven graph reasoning, while the GPO branch preserves coarse-grained feature. Finally, the two branch-specific similarity matrices are adaptively fused, while the stop-gradient fused similarity serves as a soft target to refine the pre-fusion matrices through KL-divergence alignment.
An overview of our MSGAT is shown in Fig.~\ref{fig:framework}. We first extract floating-point image and text node features and then construct spiking visual and textual graphs. MSGAT captures fine-grained structural semantics through spike-driven graph reasoning, whereas GPO preserves coarse-grained information. The resulting similarity matrices are adaptively fused, and the stop-gradient fused similarity supervises the two pre-fusion matrices via KL-divergence alignment.

\subsubsection{Spiking Neuron} 
As the fundamental unit of SNNs, a spiking neuron receives input currents $X[t]$ and accumulates membrane potentials, which are compared to a threshold to determine firing. The dynamics of the leaky integrate-and-fire (LIF) neuron model~\cite{third-network-LIF} are: 
\begin{align} 
	H[t] & = V[t-1] + \frac{1}{\tau} \left(X[t] - \left(V[t-1]-V_\mathrm{reset} \right) \right), \\ 
	S[t] & = \begin{cases} 1, & \text{if }H[t] \ge V_{\mathrm{th}}, \\ 0, & \text{otherwise}, \end{cases} \\ 
	V[t] & = H[t] \left(1 - S[t] \right) + V_\mathrm{reset}S[t].
\end{align}
where $\tau$ is the membrane time constant. When $H[t]$ exceeds the threshold $V_{\mathrm{th}}$, the neuron emits a spike $S[t]$; if spiking occurs, $V[t]$ resets to $V_\mathrm{reset}$, otherwise it remains $H[t]$.

\subsection{Spiking  Graph Construction}

Following the standard practice in ITR tasks~\cite{SCAN,VSE++,VSE-infty}, we use pre-extracted visual region features by Faster-RCNN~\cite{FasterRCNN} as model inputs: 
$
\mathbf{X}^{v}=
[\mathbf{x}^{v}_{1}, \mathbf{x}^{v}_{2}, \ldots, \mathbf{x}^{v}_{N}]^{\top}
\in \mathbb{R}^{N \times 2048},
$
and a pre-trained BERT~\cite{BERT} to extract word features:
$
\mathbf{X}^{w}=
[\mathbf{x}^{w}_{1}, \mathbf{x}^{w}_{2}, \ldots, \mathbf{x}^{w}_{L}]^{\top}
\in \mathbb{R}^{L \times 768}
$. $\mathbf{x}^{w}_{i}$ and $\mathbf{x}^{v}_{i}$ denote the embedding of the $i$-th textual node and visual node respectively. Subsequently, a linear layer maps them to a shared embedding dimension of $D$.

%This setting decouples the feature extractor from the retrieval module and has been widely adopted by ANN-based baselines~\cite{MaxMatch,VSE-infty,HREM,CHAN,USER}, enabling fair comparison between different retrieval networks. 

\subsubsection{Spiking Textual Graph}
We construct a textual graph by combining explicit syntactic dependencies and latent semantic correlations. To capture explicit linguistic structures, we use Stanford CoreNLP~\cite{CoreNLP} to extract syntactic dependency relations among textual nodes. Based on the parsed dependency relations, we construct a syntactic adjacency matrix:
$
\mathbf{A}^{w}_{sy}
\in\{0,1\}^{L \times L}.
$
Since syntactic dependencies may not cover all latent semantic correlations, we further construct a semantic graph:
\begin{align}
\mathbf{A}^{w}_{se}
&=
\mathbb{I}(cosine
(
\mathbf{x}^{w}_{i},
\mathbf{x}^{w}_{j}
)>\theta^w)\in\{0,1\}^{L \times L},
\end{align}
where $\mathbb{I}(\cdot)$ denotes the indicator function, which assigns a value of 1 if the cosine similarity $cosine(\cdot)$ between two nodes exceeds the threshold $\theta^w$, and 0 otherwise. 
Finally, the syntactic dependency matrix and semantic similarity matrix are combined as:
\begin{align}
\mathbf{A}^{w}=
\mathbb{I}((
\mathbf{A}^{w}_{sy}
+
\mathbf{A}^{w}_{se})>0)\in\{0,1\}^{L \times L}.
\end{align}

%In this way, $\mathbf{A}^{w}$ incorporates both dependency-based linguistic priors from CoreNLP and embedding-based semantic correlations from pre-trained BERT.

\subsubsection{Spiking Visual Graph}
The regions detected by Faster R-CNN may contain multiple instances of the same object category, such as several "people" within a single image. Since these instances typically exhibit similar embeddings, it is reasonable to establish edges between the corresponding nodes. Accordingly, we construct the adjacency matrix of the visual graph based on the pairwise cosine similarity:
\begin{align}
\mathbf{A}^{v}
&=
\mathbb{I}(cosine
(
\mathbf{x}^{v}_{i},
\mathbf{x}^{v}_{j}
)>\theta^v)\in\{0,1\}^{N \times N},
\end{align}
where $\theta^v$ is the threshold of visual node-level similarity.

\subsubsection{Spike Encoder}
To meet the spatiotemporal requirements of SNNs, we employ a Spike Encoder to convert floating-point  node features into spiking pattern embeddings~\cite{SeqSNN}. Given an input  floating-point tensor $X_f$, we first apply layer normalization and then replicate $T$ time steps. %To provide each time step with a distinct representation, a learnable temporal affine transformation is applied before spike generation: % 仿射变换
\begin{align}
X_{s}^{\,t}
=
\mathcal{SN}
\left(
\gamma^{t}
\odot
\mathrm{LN}\left(X_{f}\right)
+\beta^{t}\right)
,  t=1,\ldots,T.
\end{align}
where $\gamma^{t}$ and $\beta^{t}$ are learnable scale and bias parameters associated with the $t$-th time step, respectively, and $\odot$ denotes element-wise multiplication, $\mathcal{SN}(\cdot)$ is the spiking neuron.

\subsection{Multi-head Spiking Graph Attention}

Graph attention networks~\cite{GAT} provide an effective mechanism for modeling relational dependencies by adaptively aggregating information from neighboring nodes. GATv2~\cite{GATv2} improves the original attention formulation by allowing the attention ranking to dynamically vary with different source-target node pairs, enabling more expressive graph reasoning. %However, conventional graph attention relies on continuous feature interactions and Softmax function, which are incompatible with event-driven SNNs. To bridge dynamic graph attention and spike-based computation, 

Inspired by the dynamic source-target interaction in GATv2, we propose a Multi-head Spiking Graph Attention Network (MSGAT). Given spiking node features $\mathbf{X}_s\in\{0,1\}^{T\times N\times D}$, MSGAT first generates spiking source ($\mathbf{Q}$), target ($\mathbf{K}$), and value ($\mathbf{V}$) matrix though linear projections and LIF neuron respectively:
\begin{align}
(\mathbf{Q},\mathbf{K},\mathbf{V})
=
\operatorname{LIF}
\left(
\operatorname{BN}
\left(
\mathbf{X}_s
(\mathbf{W}_q,\mathbf{W}_k,\mathbf{W}_v)
\right)
\right),
\end{align}
where $\mathbf{W}_q,\mathbf{W}_k$ and $\mathbf{W}_v$ are projection weights. Then, multi-head attention mechanism is adopted to split the spike tensor into $H$ sub-space. For the $h$-th head, they are reshaped as:
\begin{align}
\mathbf{q}^{h},
\mathbf{k}^{h},
\mathbf{v}^{h}
&\in
\{0,1\}^{T\times N\times d},
&
d
&=D/H.
\end{align}

To compute dynamic source-target attention, MSGAT first constructs a vector-valued edge response for each pair of nodes $i$ and $j$. The response for  head $h$ is defined as:
\begin{align}
e_{ij}^{h}
&= \frac{1}{T}\sum_{t=1}^T\left(
{\mathbf{a}_{q}^{h}} \odot
\mathbf{q}_{i}^{h}[t]
+
{\mathbf{a}_{k}^{h}} \odot
\mathbf{k}_{j}^{h}[t]\right)
+
b^{h}\mathbf{1}{d},
\end{align}
where $\mathbf{a}_{q}^{h}$ and $\mathbf{a}_{k}^{h}$ are learnable head-specific attention vectors, and $b^{h}$ is a learnable broadcast scalar bias, $\odot$ denotes Hadamard product. Since $\mathbf{q}_{i}^{h}[t]$and $\mathbf{k}_{j}^{h}[t]$ are binary, their element-wise products with the attention vectors can be computed by accumulating the corresponding weights. 

MSGAT applies ReLU  before feature reduction, allowing target attention rankings to vary across source nodes, and subsequently incorporates the binary adjacent matrix $\mathbf{A} \in \{0,1\}^{N \times N}$ to preserve only valid graph neighborhoods:
\begin{align}
\bar{e}_{ij}^{h}
&=
\mathbf{A}_{ij} \odot
\sum_{r=1}^d
\operatorname{ReLU}
\left(
e_{ij,r}^{h}
\right) \in \mathbb{R}_{\ge 0}^{h \times N \times N}
,
\end{align}
where ReLU($\cdot$) can be implemented by simple thresholding operations in neuromorphic systems.

To avoid the exponential computation introduced by SoftMax, MSGAT adopts row-wise normalization:
\begin{align}
\tilde{e}_{ij}^{h}
&=
\frac{
\bar e_{ij}^{h}
}
{
\sum_{j\in\mathcal N(i)}
\bar e_{ij}^{h}
+\epsilon
} \in [0,1)^{h \times N \times N},
\end{align}
where $\epsilon$ is a small constant that prevents numerical instability when the denominator becomes zero due to sparse responses. The spike-driven  neighborhood aggregation is formulated as:
\begin{align}
\mathbf{o}^{h}
&=
\sum_{j=1}^{N}
\tilde{e}_{ij}^{h}
\mathbf{v}_{j}^{h} \in
[0,1)^{T\times h \times N\times d}.
\end{align}

Finally, the complementary information from all source-target dynamic attention heads is concatenated as:
\begin{align}
\mathbf{O}
&=
\mathbf{o}^{1} \Vert
\mathbf{o}^{2} \Vert
\ldots \Vert \mathbf{o}^{H}
\in
[0,1)^{T\times N\times D}.
\end{align}
where "$\Vert$" indicates concatenation. The aggregated spiking representation is further converted into spike outputs through a LIF neuron:
$\hat{\mathbf{O}}
=
\operatorname{LIF}
\left(
\mathbf{O} 
\right) \in
\{0,1\}^{T\times N\times D}$.

%In summary, MSGAT provides a spike-compatible reformulation of dynamic graph attention. It inherits the source-target dependency modeling of GATv2, while replacing softmax-based continuous attention with non-negative edge responses,adjacent matrix masking, and spike-driven message aggregation. Moreover, it avoids expensive continuous attention normalization, making it suitable for efficient spike-based computation.

\subsection{Similarity-Space Fuse Alignment Strategy}

As illustrated in Fig.~\ref{fig:motivation}, fine-grained  representations from the MSGAT exhibit sparse and discrete characteristics, which are misaligned with the continuous GPO representations. This representation discrepancy makes feature-level fusion suboptimal, as verified in Fig.~\ref{fig:fuse-menthods}. To address this issue, we propose Similarity-Space Fuse Alignment Strategy (Sim-Fuse), which bypasses representation-level fusion and performs multi-granularity alignment  in the similarity space. 

For fine-grained branch, we compute the fine-grained similarity tensor: $\mathbf{s}(\mathbf{X}^v,\mathbf{X}^w)\in\mathbb{R}^{B\times B\times N\times L}$, $B$ denotes the batch size. Then, we aggregate similarities from the word-to-region direction by selecting the most relevant region for each word~\cite{CHAN}, and the most relevant word for each region along the  reverse direction:
\begin{align}
    \mathbf{s}_{w\rightarrow r}
=
\frac{1}{L}
\sum_{l=1}^{L}
\max_{n}
\mathbf{s}_{n,l},  \; \; 
\mathbf{s}_{r\rightarrow w}
=
\frac{1}{N}
\sum_{n=1}^{N}
\max_{l}
\mathbf{s}_{n,l}.
\end{align}
Finally, we combine the fine-grained similarities, defined as:
\begin{align}
   \mathbf{S}_f
=
\frac{
\mathbf{s}_{w\rightarrow r}
+
\mathbf{s}_{r\rightarrow w}
}{2}
\in
\mathbb{R}^{B\times B}.
\label{eq: sime-fine}
\end{align}

For coarse-grained branch, we use Generalized Pooling Operator (GPO) to aggregate region and word nodes into global image and text embeddings: $\mathbf{g}^{v}$ and $\mathbf{g}^{w} \in \mathbb{R}^{B \times D}$ , and the coarse-grained similarity is computed as:
$
\mathbf{S}_{c}=cosine(\mathbf{g}^{v},\mathbf{g}^{w}).
$
The fused similarity matrix $\mathbf{S}_{fuse}$is obtained by a learnable weighted combination of $\mathbf{S}_{f}$ and $\mathbf{S}_{c}$:
\begin{equation}
\mathbf{S}_{fuse}=\lambda_{f}\mathbf{S}_{f}+\lambda_{c}\mathbf{S}_{c} .
\label{eq:sim-fuse}
\end{equation}
Next, each row and column of the similarity matrices is converted into matching probability distribution $\mathbf{R}$ and $\mathbf{C}$ using temperature-scaled softmax. The image-to-text alignment loss is defined as:
\begin{equation}
\mathcal{L}_{KL}^{i2t}=D_{KL}(\bar{\mathbf{R}}_{fuse}\Vert\mathbf{R}_{f})+D_{KL}(\bar{\mathbf{R}}_{fuse}\Vert\mathbf{R}_{c}),
\end{equation}
where $\bar{\quad}$ denotes the stop-gradient operation. Similarly, the text-to-image alignment loss is defined as:
\begin{equation}
\mathcal{L}_{KL}^{t2i}=D_{KL}(\bar{\mathbf{C}}_{fuse}\Vert\mathbf{C}_{f})+D_{KL}(\bar{\mathbf{C}}_{fuse}\Vert\mathbf{C}_{c}).
\end{equation}
The final similarity alignment loss is:
\begin{equation}
\mathcal{L}_{KL}=\frac{1}{2}\left(\mathcal{L}_{KL}^{i2t}+\mathcal{L}_{KL}^{t2i}\right).
\label{eq:loss-kl}
\end{equation}

Through this objective, $\mathbf{S}_{fuse}$ integrates complementary coarse- and fine-grained matching cues and provides guidance for optimizing $\mathbf{S}_{f}$ and $\mathbf{S}_{c}$. In turn, the refined branch similarities further enhance the fused similarity, forming a collaborative refinement process. 
\subsection{Training Objectives}
We improve the  hubness-aware loss~\cite{hub-loss} as our training objective. For the similarity matrix $\textbf{S}$, we define the $\textbf{S}_{ii}, i \in [1,B]$ as positive pairs, $\textbf{S}_{ij}$ and $\textbf{S}_{ji}, j \in [1,i) \cup (i,B]$ as negative pairs. The loss function is formulated as:
\begin{align}
\mathcal{L}
&=
\frac{1}{B}
\sum_{i=1}^{B}
(
\frac{1}{\mu}
\log
(1+\sum_{j=1,j \ne i}^{B} e^{\mu(\bar{\mathbf{S}}_{ij}-\gamma)}) \notag\\
&+ 
\frac{1}{\mu}
\log
(1+\sum_{j=1, j\ne i}^{B} e^{\mu(\bar{\mathbf{S}}_{ji}-\gamma)}) 
-\log(1+\mathbf{S}_{ii})
),
\label{eq:loss-function}
\end{align}
where $\mu$  and $\gamma$ are temperature scale and margin,
respectively.

\subsubsection{Node-level Objective}
Sparse spike representations are sensitive to noise, and unreliable activations introduced at early stages may be propagated and amplified during subsequent spiking graph reasoning. To stabilize the inputs  of MSGAT, we impose an auxiliary node-level objective before Spike Encoder. Given the visual nodes  $X^v$ and textual nodes $X^w$, the similarity $\mathbf{S}_{node}$ is computed by Eq.\ref{eq: sime-fine}, and according to Eq.\ref{eq:loss-function}, 
$
\mathcal{L}_{n}
=
\mathcal{L}
\left(
\mathbf{S}_{node}
\right).
$
\subsubsection{Structure-level Objective}
After MSGAT reasoning, each node aggregates information from its graph neighborhood and encodes relation-enhanced structural semantics. The structure-level objective is defined as 
$
\mathcal{L}_{s}
=
\mathcal{L}\left(\mathbf{S}_{f} \right)
+
\mathcal{L}\left(\mathbf{S}_{c} \right)
$
to capture local and global structural information.

According to Eq.\ref{eq:sim-fuse}, the loss function of fused similarity matrix is defined as:
$
  \mathcal{L}_{f}
=
\mathcal{L}
\left(
\mathbf{S}_{fuse}
\right)
$. It encourages the model to jointly optimize fine-grained branch  and coarse-grained branch with learnable weights. And the final total objective is formulated as:
\begin{align}
    \mathcal{L}_{total} =\mathcal{L}_f + \lambda_1 \mathcal{L}_n + \lambda_2 \mathcal{L}_s + \lambda_3 \mathcal{L}_{KL}
\end{align}
where $\lambda_1$, $\lambda_2$ and $\lambda_3$ are weights to balance different losses.
\section{Experiments}
\subsection{Datasets Metrics and Settings}
We evaluate MSGAT on two widely used benchmarks: \textsc{Flickr30K}~\cite{Flickr30K} and \textsc{MSCOCO}~\cite{MSCOCO}. \textsc{Flickr30K} contains 31,783 images and \textsc{MSCOCO} includes 123,287 images, each paired with five captions. Following the standard evaluation protocol~\cite{SCAN,USER}, \textsc{Flickr30K} is split into 29,000/1,000/1,000 images for training/validation/testing, while \textsc{MSCOCO} uses 113,287/1,000/5,000 images. 

The commonly used evaluation metrics for ITR are Recall@K (K=1,5,10), which depict the percentage of ground truth being retrieved at top 1, 5, 10 results, respectively. To show the overall matching performance, we also
compute the sum of all the Recall values (R@Sum). %at image-to-text and text-to-image directions.

For a fair comparison, we employ Faster R-CNN and BERT as feature extractors. MSGAT is implemented with the SpikingJelly~\cite{Spikingjelly} framework and trained on a single NVIDIA RTX4090 GPU. More parameter and model setting details are provided in the Supplementary Materials.

\subsection{Main  Experimental Results}

We compare MSGAT with state-of-the-art ANN-based methods and the existing SNN retrieval baseline on Flickr30K and MSCOCO, as shown in Tab.~\ref{tab:all-results} and Tab.~\ref{tab:mscoco5k}. Methods are grouped according to their feature extraction backbones, and unlike SCAN~\cite{SCAN}, VSRN~\cite{VSRN},  and VSRN++~\cite{VSRN++}, which boost performance via multi-model ensembling, MSGAT reports single-model results like MMCA~\cite{MMCA} and NAAF~\cite{NAAF}. On the \textsc{MSCOCO} 1K (5-fold) and \textsc{Flickr30K} 1K test sets, MSGAT achieves R@Sum scores of 539.0 and 529.8, respectively, outperforming all compared methods. Compared with CMSF, MSGAT improves R@Sum by 5.2 and 5.9 points on the two benchmarks, demonstrating the effectiveness of spike-driven graph reasoning and similarity-space fusion. Moreover, with only two time steps, MSGAT surpasses its ANN counterpart by 2.0 and 2.6 R@Sum points, showing the potential of SNNs for efficient cross-modal retrieval.
On the more challenging \textsc{MSCOCO} 5K test set (Tab.~2), MSGAT achieves the best R@Sum of 452.0 and consistently outperforms its ANN counterpart across most metrics with T=2. Although MaxMatch~\cite{MaxMatch} achieves competitive performance,  %through explicit matching optimization, 
it relies on Hungarian-based matching~\cite{hungarian} with additional computational complexity.  In contrast, MSGAT provides efficient structure-aware reasoning and similarity-level alignment under sparse spike computation. Compared with the SNN baseline CMSF~\cite{CMSF}, MSGAT is slightly lower on text-to-image R@5 and R@10, but achieves much better overall performance with a simpler architecture and training pipeline.
\begin{table*}[!t]
	\centering
	\footnotesize
	\setlength{\tabcolsep}{1.5mm}
	
	\begin{tabular}{l|l|l|p{0.6cm}p{0.6cm}p{0.7cm}|p{0.6cm}p{0.6cm}p{0.7cm}|c|p{0.6cm}p{0.6cm}p{0.7cm}|p{0.6cm}p{0.6cm}p{0.7cm}|c}
    %{l|l|l|ccc|ccc|c|ccc|ccc|c}

	\toprule[1.1pt]
	\multirow{3}[4]{*}{Method}
	& \multirow{3}[4]{*}{\makecell{Venue}}
    & \multirow{3}[4]{*}{\makecell{S}}
	& \multicolumn{7}{c|}{\textsc{MSCOCO} 1K (5-fold) Test Set}
	& \multicolumn{7}{c}{\textsc{Flickr30K} 1K Test Set} \\ 
	
	\cmidrule(lr){4-10}
	\cmidrule(rl){11-17}
	
	& &
	& \multicolumn{3}{c|}{Image-to-Text}
	& \multicolumn{3}{c|}{Text-to-Image}
	& \multirow{2}[2]{*}{\makecell{R@\\Sum}}
	& \multicolumn{3}{c|}{Image-to-Text}
	& \multicolumn{3}{c|}{Text-to-Image}
	& \multirow{2}[2]{*}{\makecell{R@\\Sum}} \\ 
	\cmidrule(lr){4-6}
	\cmidrule(lr){7-9}
	\cmidrule(lr){11-13}
	\cmidrule(lr){14-16}
	&&& R@1 & R@5 & R@10& R@1 & R@5 & R@10
	&& R@1 & R@5 & R@10& R@1 & R@5 & R@10
	& \\
	
	\midrule
	\multicolumn{17}{c}{\textit{Faster R-CNN + BiGRU}} \\ 
	\midrule
	SCAN$\ast$ &ECCV18&& 72.7 & 94.8 & 98.4 & 58.8 & 88.4 & 94.8 & 507.9 & 67.4 & 90.3 & 95.8 & 48.6 & 77.7 & 85.2 & 465.0 \\ 
	VSRN$\ast$ &ICCV19&& 76.2 & 94.8 & 98.2 & 62.8 & 89.7 & 95.1 & 516.8 & 71.3 & 90.6 & 96.0 & 54.7 & 81.8 & 88.2 & 482.6 \\ 
	VSE$\infty$ &CVPR21&& 78.5 & 96.0 & 98.7 & 61.7 & 90.3 & 95.6 & 520.8 & 76.5 & 94.2 & 97.7 & 56.4 & 83.4 & 89.9 & 498.1 \\ 
	NAAF~ &CVPR22&& 78.1 & 96.1 & 98.6 & 63.5 & 89.6 & 95.3 & 521.2 & 79.6 & 96.3 & 98.3 & 59.3 & 83.9 & 90.2 & 507.6 \\ 
	CMSF & CVPR26&\checkmark&78.9 & 96.2 & 98.7	& 63.6	& 91.0	& 96.3	& 524.7 & 80.7	& 95.0	& 97.6	& 61.3	& 85.9	& 91.3	& 511.8\\
    \rowcolor{gray!20}
    MSGAT&Ours (T=2)&\checkmark& 80.5& 96.5& 98.4& 64.4& 91.0& 96.2& 527.0&82.2& 95.6& 97.9& 61.8& 86.5& 92.5& 516.5\\
    
	\midrule
	\multicolumn{17}{c}{\textit{Faster R-CNN + BERT}} \\ 
	\midrule
	MMCA &CVPR20&& 74.8 & 95.6 & 97.7 & 61.6 & 89.8 & 95.2 & 514.7 & 74.2 & 92.8 & 96.4  &54.8 & 81.4 & 87.8 & 487.4 \\ 
	%TERAN$\ast$ &&& 80.2 & 96.6 & \textbf{99.0} & \underline{67.0} & \underline{92.2} & \textbf{96.9} & 531.9 & 79.2 & 94.4 & 96.8 & 63.1 & 87.3 & 92.6 & 513.4 \\ 
	VSE$\infty$ &CVPR21&& 79.7 & 96.4 & 98.9 & 64.8 & 91.4 & 96.3 & 527.5 & 81.7 & 95.4 & 97.6 & 61.4 & 85.9 & 91.5 & 513.5 \\ 
	VSRN++$\ast$~ &TPAMI23& &77.9 & 96.0 & 98.5 & 64.1 & 91.0 & 96.1 & 523.6 & 79.2 & 94.6 & 97.5 & 60.6 & 85.6 & 91.4 & 508.9 \\ 
	CHAN &CVPR23 &&81.4 & 96.9 & 98.9 & 66.5 & 92.1 & \underline{96.7} & 532.6 & 80.6 & 96.1 & 97.8 & 63.9& 87.5 & 92.6 & 518.5 \\ 
	HREM &CVPR23&& 81.1 & 96.6 & 98.9 & 66.1 & 91.6 & 96.5 & 530.7 & 83.3 & 96.0 & 98.1& 63.5 & 87.1 & 92.4 & 520.4 \\ 
	USER &TIP24&& 82.8 & 96.8 & 98.8 & 66.1 & 90.6 & 95.6 & 530.5 & 82.7 & 97.0 & 98.3 & 63.1 & 86.7 & 92.1 & 519.9 \\ 
	
    CORA &CVPR24&&82.4 &96.8& 98.8 &66.2& 91.9& 96.6 &532.7& 83.7 &96.6 &98.3 &62.3 &87.1 &92.6 &520.1\\
    MaxMatch &ACL25&& 83.0 & 96.9 & 98.9 & 66.4 & 91.9 & 96.6 & 533.8 & 84.2 & 96.1 & 97.9 & 63.2 & 87.3 & 92.2 & 520.8 \\ 
	CMSF & CVPR26&\checkmark& 81.9 & 96.7 & \underline{98.9} & 67.1 & 92.4 & 96.8 & 533.8 & 82.1 & 96.3 & 98.0 & 65.9 & \textbf{88.4} & \textbf{93.2} & 523.9 \\
    \rowcolor{gray!20}
    MSGAT &Ours (ANN)&&\underline{84.8}&
\underline{97.1}&
98.8&
\underline{67.4}&
\underline{92.4}&
96.5&
\underline{537.0}&
\underline{85.7}&
\underline{97.3}&
\underline{98.7}&
\underline{65.4}&
87.4&
\underline{92.7}&
\underline{527.2}\\
\rowcolor{gray!20}
    MSGAT&Ours (T=2)&\checkmark&
\textbf{85.8}&
\textbf{97.1}&
\textbf{99.0}&
\textbf{67.8}&
\textbf{92.5}&
\textbf{96.8}&
\textbf{539.0}&
\textbf{86.1}&
\textbf{98.4}&
\textbf{99.0}&
\textbf{65.9}&
\underline{87.8}&
92.6&
\textbf{529.8} \\
	\bottomrule[1.1pt]
	\end{tabular}
    \caption{Results on the \textsc{MSCOCO } and \textsc{Flickr30K 1K} test sets. Best and second-best results are marked in bold and \underline{underline}, respectively. $\ast$ denotes ensemble results, and S indicates spike-driven methods. For fair comparison, we report only Faster R-CNN- and BERT/BiGRU-based baselines; results with pretrained VLMs are provided in the Supplementary Material.}
	\label{tab:all-results}
\end{table*}

\begin{table}[!t]
	\footnotesize
	\centering
	\setlength{\tabcolsep}{3pt}
	
	\begin{tabular}{l|ccc|ccc|c}
	\toprule[1.1pt]
	\multirow{3}[4]{*}{Methods} & \multicolumn{7}{c}{\textsc{MSCOCO} 5K Test Set} \\ 
	\cmidrule(lr){2-8}
	& \multicolumn{3}{c|}{Image-to-Text} & \multicolumn{3}{c|}{Text-to-Image} & \multirow{2}[2]{*}{\makecell{R@\\Sum}} \\ 
	\cmidrule(lr){2-4} \cmidrule(lr){5-7}
	& R@1 & R@5 & R@10 & R@1 & R@5 & R@10 & \\ 
	\midrule
	\multicolumn{8}{c}{\textit{Faster R-CNN + BiGRU}} \\ 
	\midrule
	SCAN$\ast$ & 50.4 & 82.2 & 90.0 & 38.6 & 69.3 & 80.4 & 410.9 \\ 
	VSRN$\ast$ & 53.0 & 81.1 & 89.4 & 40.5 & 70.6 & 81.1 & 415.7 \\ 
	VSE$\infty$ & 56.6 & 83.6 & 91.4 & 39.3 & 69.9 & 81.1 & 421.9 \\ 
	NAAF & 58.9 & 85.2 & 92.0 & 42.5 & 70.9 & 81.4 & 430.9 \\ 
    CMSF  & 58.5	& 85.3	& 92.5	& 42.5	& 72.0	& 82.2	& 432.0 \\
	\rowcolor{gray!20}
	MSGAT$_{\mathrm{T=2}}$ &60.4& 85.6& 92.6& 43.4& 72.9& 83.4& 438.3\\
    
	\midrule
	\multicolumn{8}{c}{\textit{Faster R-CNN + BERT}} \\ 
	\midrule
	MMCA & 54.0 & 82.5 & 90.7 & 38.7 & 69.7 & 80.8 & 416.4 \\ 
	%TERAN$\ast$ & 59.3 & 85.8 & 92.4 & 45.1 & \textbf{76.4} & \underline{84.4} & 443.4 \\ 
	VSE$\infty$& 58.3 & 85.3 & 92.3 & 42.4 & 72.7 & 83.2 & 434.2 \\ 
	VSRN++$\ast$ & 54.7 & 82.9 & 90.9 & 42.0 & 72.2 & 82.7 & 425.4 \\ 
	CHAN~ & 59.8 & 87.2 & 93.3 & 44.9 & 74.5 & 84.2 & 443.9 \\ 
	HREM & 62.3 & 87.6 & 93.4 & 43.9 & 73.6 & 83.3 & 444.1 \\ 
	USER & 63.7& 87.4 & 93.5 & 44.8 & 73.4 & 82.7 & 445.5 \\ 
	 
    CORA &62.4 &86.8 &92.6& 44.2& 73.6& 83.9 &443.6\\
    MaxMatch & 63.3 & 87.9 & 93.2 & 44.2 & 73.9 & 83.9 & 446.5 \\
	CMSF & 61.5 & 86.7 & 92.8 & 45.1 & \textbf{75.0} & \textbf{84.6} & 445.8 \\ 
    
    \rowcolor{gray!20}
    MSGAT$_{\mathrm{ANN}}$ &\underline{64.7} &\underline{87.9} &\underline{93.5}& \underline{45.1} &73.9& 83.6 &\underline{448.8}\\
    \rowcolor{gray!20}
    MSGAT$_{\mathrm{T=2}}$ & \textbf{65.2} &\textbf{88.7} &\textbf{93.9} &\textbf{45.5} &\underline{74.5} &\underline{84.3} &\textbf{452.0}\\
	\bottomrule[1.1pt]
	\end{tabular}
    \caption{Results on \textsc{MSCOCO} 5K Test Set. Best results are in bold, and the second-best scores are \underline{underlined}. $T=2$ indicates the number of time steps in MSGAT, while ANN denotes the ANN counterpart of MSGAT.}
	\label{tab:mscoco5k}
\end{table}

\subsubsection{Efficiency Analysis of MSGAT}
\begin{figure}[!t]
	\centering
	\includegraphics[width = \linewidth]{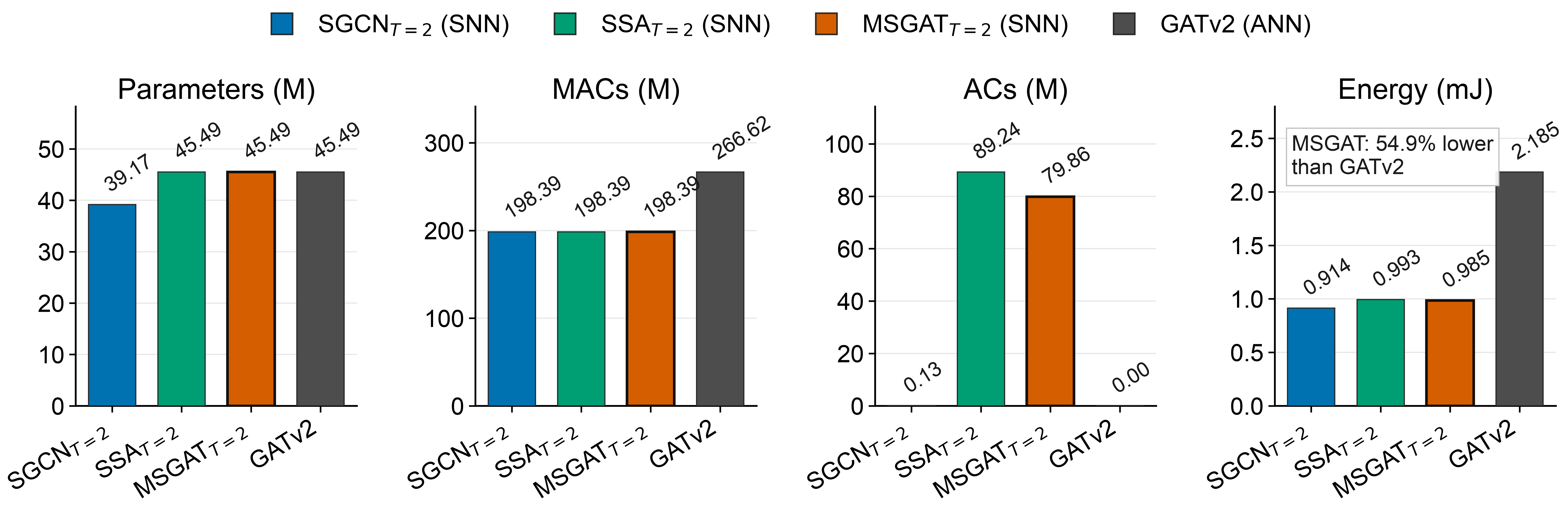} % 图像文件名
	\caption{Efficiency Analysis. SSA: spike self-attention; AC: accumulate; MAC: multiply-accumulate operation.}
	\label{fig:energy}
\end{figure}
As shown in Fig.~\ref{fig:energy}, we compare MSGAT with other methods in terms of parameters, MACs, ACs, measured by \cite{syops}, and theoretical inference energy consumption estimated following \cite{SeqSNN}. The reported cost excludes the replaceable feature extractors, i.e., Faster R-CNN and BERT. SSA, MSGAT, and GATv2 have the same number of parameters, 45.49M, since they all use four linear projection layers in attention block. For SNN-based methods, the MACs mainly come from projecting floating-point features into the same dimension $D$, leading SGCN, SSA, and MSGAT to share the same MACs cost of 198.39M. For ACs, SGCN only requires 0.13M because its weight tensor passes through the neuron once simply, while MSGAT also reduces 10.5\% ACs compared with SSA. Overall, MSGAT consumes only 0.985 mJ, reducing theoretical inference energy by 54.9\% compared with GATv2 and demonstrating strong energy efficiency.

\subsection{Ablation Studies}
\subsubsection{Ablation on MSGAT Module}
Tab.~\ref{tab:fine-grained branch} evaluates different fine-grained branch designs. Removing this branch gives an R@Sum of 512.7, confirming its importance. SSA and SGCN improve performance to 520.4 and 524.7, respectively; however, SGCN cannot model dynamic attention among nodes, while SSA may be disturbed by irrelevant node interactions.
MSGAT achieves the best R@Sum of 529.8, outperforming w/o fine-grained matching, SSA, and SGCN by 17.1, 9.4, and 5.1 points, respectively. It also performs best on most recall metrics and surpasses the GAT-ANN counterpart by 2.6 R@Sum. These results show that MSGAT provides stronger structure-aware fine-grained matching under sparse spike representations. We further attribute this advantage partly to Sim-Fuse, which is better suited to heterogeneous features with different distributions than continuous ANN features.

\subsubsection{Ablation on Sim-Fuse Strategy}
\begin{figure}[!t]
	\centering
	\includegraphics[width = \linewidth]{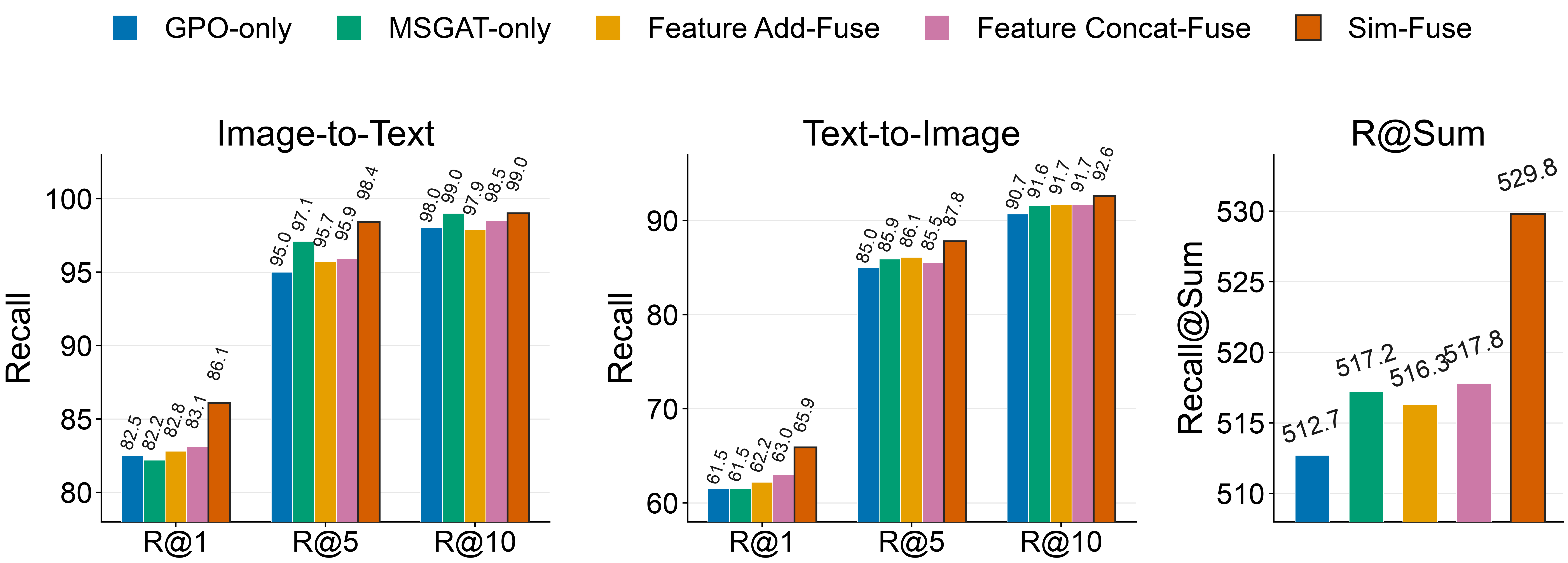} % 图像文件名
	\caption{Effectiveness of Similarity-Space Fuse strategy on \textsc{Flickr30K} 1K test set. Add-Fuse and Concat-Fuse denote features addition and concatenation, respectively.}
	\label{fig:fuse-menthods}
\end{figure}

Fig.~\ref{fig:fuse-menthods} compares coarse- and fine-grained fusion strategies on the Flickr30K 1K test set. GPO-only and MSGAT-only achieve R@Sum scores of 512.7 and 517.2, respectively, demonstrating the benefit of structure-aware fine-grained matching. However, feature-level fusion provides limited gains: Add-Fuse decreases R@Sum to 516.3 due to interference between discrete MSGAT and continuous GPO representations, while Concat-Fuse reaches only 517.4, with its marginal improvement potentially arising from the doubled feature dimensionality rather than effective cross-granularity interaction. In contrast, Sim-Fuse bypasses this feature-space mismatch and integrates both branches directly in the similarity space, substantially improving R@Sum to 529.8.

\subsubsection{Ablation on Training Objectives}
Tab.~\ref{tab:loss-function} investigates the effect of different training objectives. Using only the  fused loss $\mathcal{L}_{f}$ achieves an R@Sum of 522.2, while adding the node-level objective $\mathcal{L}_{n}$ further improves it to 525.0 by providing stable spike representations before MSGAT reasoning. Since $\mathbf{S}_{fuse}$ is derived from fine- and coarse-grained similarities $\mathbf{S}_{f}$ and $\mathbf{S}_{c}$ , introducing structure-level supervision $\mathcal{L}{s}$ enhances individual similarity  and increases R@Sum to 526.4. Finally, $\mathcal{L}_{KL}$ aligns $\mathbf{S}_{f}$ and $\mathbf{S}_{c}$ with the detached fused similarity distribution, achieving the best performance of 529.8. These results verify the effectiveness of each objective.

\subsubsection{Ablation on Time Steps}
\begin{figure}[!t]
	\centering
	\includegraphics[width = \linewidth]{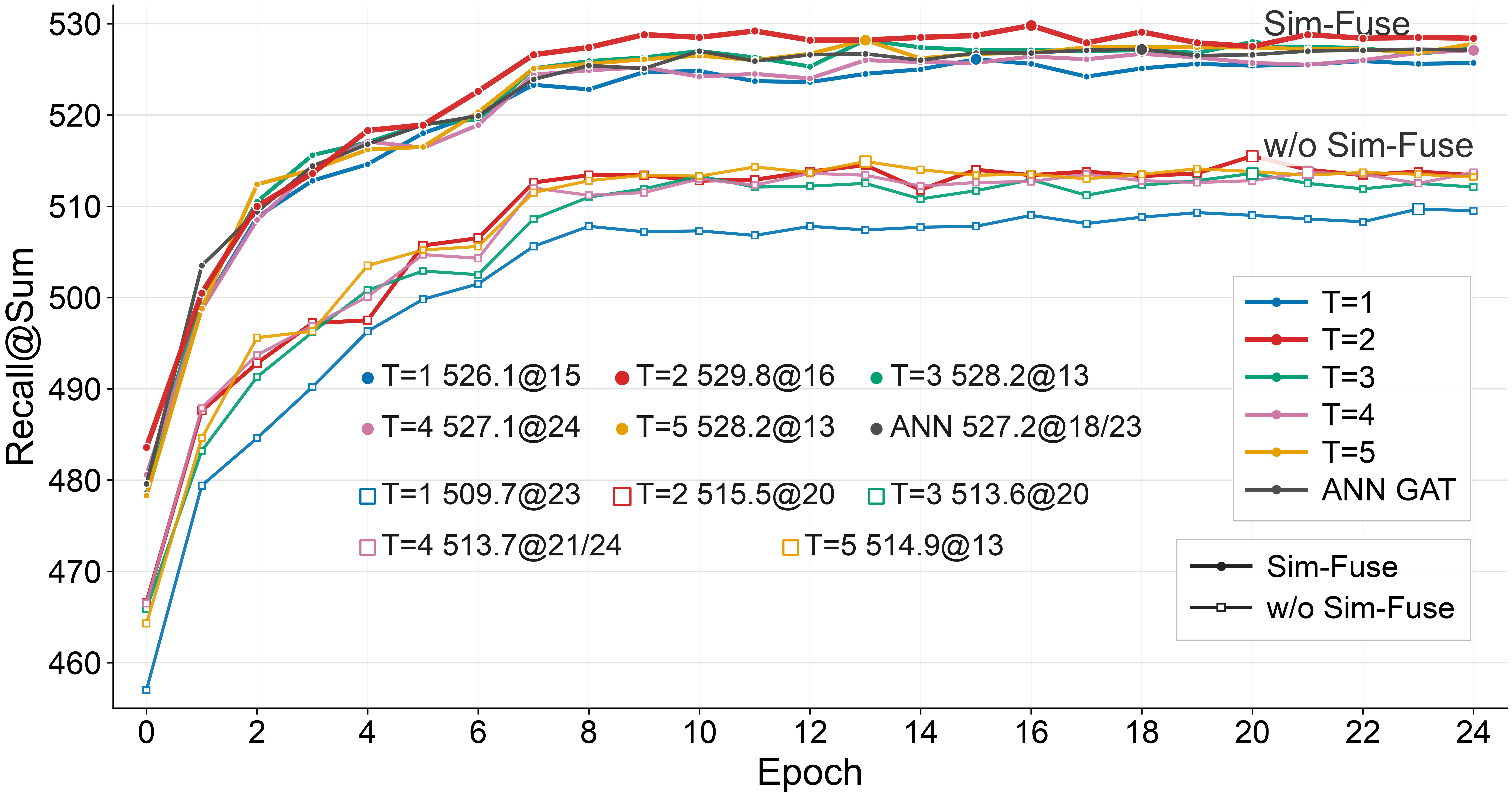} % 图像文件名
	\caption{ Training Performance of MSGAT with 1 to 5 Time Steps. The upper set of curves corresponds to  Sim-Fuse, whereas the lower set corresponds to without Sim-Fuse.}
	\label{fig:timestep}
\end{figure}
The time steps  ablation in Fig.~\ref{fig:timestep} shows that MSGAT achieves the best performance at T=2, reaching a R@Sum of 529.8. This result outperforms the ANN-based GATv2 and other spiking variants, indicating that a small number of time steps is sufficient for effective spiking graph representation learning. Although increasing the time step beyond T=2 does not further improve the best result, it shows that MSGAT is robust to different temporal configurations. Without Sim-Fuse, the performance at T=1 decreases substantially to 509.7, suggesting that Sim-Fuse plays an important role in stabilizing training. Overall, T=2 provides the best balance between  accuracy and  efficiency.

\subsubsection{Ablation on Attention Heads}
%Table 5 analyzes the effect of attention heads in MSGAT. Multi-head decomposition enables complementary structural modeling across feature subspaces, with 8 heads achieving the best R@Sum of 529.8. More heads reduce subspace dimensionality and effective spike activations, weakening per-head representation capacity, whereas fewer heads limit subspace diversity and complementarity. Notably, with only 2 heads, the spike attention output collapses to all-zero activations, preventing effective neighborhood aggregation and graph propagation. These results indicate that 8 heads provide the best balance between subspace diversity and per-head spike representation capacity.
Tab.~\ref{tab:attention heads} analyzes the effect of the number of attention heads in MSGAT. Multi-head decomposition enables  complementary structural modeling across feature subspaces, with 8 heads achieving the best R@Sum of 529.8. More heads reduce subspace dimensionality and effective spike activations, weakening per-head representation capacity, whereas fewer heads limit subspace diversity and complementary information. %Notably, with only 2 heads, the spike attention output collapses to all-zero activations, preventing effective neighborhood aggregation and graph information propagation.
In particular, the 2-head setting degenerates into all-zero spike attention outputs after thresholding, preventing effective neighborhood aggregation and graph information propagation.
These results indicate that 8 heads provide the most stable trade-off between subspace diversity and per-head spike representation capacity.

Additional sensitivity analyses of the thresholds $\theta^v$ and $\theta^w$, ablations on Spike Encoder and the energy  estimation formula, etc., are presented in the Supplementary Material.

\begin{table}[!t]
	\centering
	\footnotesize
	\setlength{\tabcolsep}{3pt}
	
	\begin{tabular}{c|ccc|ccc|c}
	\toprule[1.1pt]
	\multirow{2}[2]{*}{\makecell{Fine-grained \\ Branch}} & \multicolumn{3}{c|}{Image-to-Text} & \multicolumn{3}{c|}{Text-to-Image} & \multirow{2}[2]{*}{\makecell{R@\\Sum}} \\ 
	\cmidrule(lr){2-4} \cmidrule(lr){5-7}
	& R@1 & R@5 & R@10 & R@1 & R@5 & R@10 & \\ 
	\midrule
	w/o & 82.5& 95.0& 98.0 & 61.5& 85.0& 90.7 & 512.7 \\ 
    SSA$_{T=2}$ & 81.9& 96.7& 98.5 & 64.1& 87.2& 92.1 & 520.4 \\
	SGCN$_{T=2}$ & 84.1& 97.4& 99.0 & 64.8& 87.0& 92.3 & 524.7 \\ 
	
    \rowcolor{gray!20}
    MSGAT$_{T=2}$ & \textbf{86.1}& \textbf{98.4}& \textbf{99.0}&\textbf{65.9}& \textbf{87.8}& 92.6&\textbf{529.8} \\ 
	\midrule
	GATv2-ANN & 85.7& 97.3&98.7& 65.4& 87.4&\textbf{92.7}&527.2\\ 
	\bottomrule[1.1pt]
	\end{tabular}
    \caption{Ablation of  fine-grained branch on \textsc{Flickr30K} 1K test set. Best results are in bold.}
	\label{tab:fine-grained branch}
\end{table}
\begin{table}[!t]
	\centering
	\footnotesize
	\setlength{\tabcolsep}{3pt}
	
	\begin{tabular}{p{0.3cm}p{0.3cm}p{0.3cm}p{0.5cm}|p{0.5cm} p{0.5cm} c|p{0.5cm} p{0.5cm}c|c}
	\toprule[1.1pt]
	 \multicolumn{4}{c|}{Loss function}& \multicolumn{3}{c|}{Image-to-Text} & \multicolumn{3}{c|}{Text-to-Image} & \multirow{2}[2]{*}{\makecell{R@ \\Sum}} \\ 
	\cmidrule(lr){1-4} \cmidrule(lr){5-7} \cmidrule(lr){8-10}
	$\mathcal{L}_f$ & $\mathcal{L}_n$ &  $\mathcal{L}_s$ &$\mathcal{L}_{KL}$& R@1 & R@5 & R@10 & R@1 & R@5 & R@10 & \\ 
	\midrule
	\checkmark &  &   && 84.1& 97.1& 98.5 & 64.2& 86.3& 92.0 & 522.2 \\ 
    \checkmark &\checkmark & & & 84.2& 97.5& 99.0& 64.9& 86.9& 92.4 & 525.0\\

	\checkmark &\checkmark & \checkmark&  & 85.6 & 97.8& 99.0&64.8& 87.0& 92.2&526.4\\ 
    
    \checkmark &\checkmark &\checkmark &\checkmark  & \textbf{86.1} & \textbf{98.4} & \textbf{99.0}& \textbf{65.9} & \textbf{87.8} & \textbf{92.6} &\textbf{529.8} \\
    
	\bottomrule[1.1pt]
	\end{tabular}
    \caption{Influence of loss function on \textsc{Flickr30K} 1K test set. Best results are in bold.}
	\label{tab:loss-function}
\end{table}
\begin{table}[!t]
	\centering
	\footnotesize
	\setlength{\tabcolsep}{3pt}
	
	\begin{tabular}{c|ccc|ccc|c}
	\toprule[1.1pt]
	\multirow{2}[2]{*}{Heads} & \multicolumn{3}{c|}{Image-to-Text} & \multicolumn{3}{c|}{Text-to-Image} & \multirow{2}[2]{*}{\makecell{R@\\Sum}} \\ 
	\cmidrule(lr){2-4} \cmidrule(lr){5-7}
	& R@1 & R@5 & R@10 & R@1 & R@5 & R@10 & \\ 
	\midrule
	2 & - & - & - & - & - & - & - \\ 
    4 & 83.6& 97.0& 98.2 & 64.6& 86.4& 92.5 & 522.3 \\ 
    
    8 & \textbf{86.1} & \textbf{98.4} & \textbf{99.0} &\textbf{65.9} & \textbf{87.8} & \textbf{92.6} &\textbf{529.8} \\ 
    16 & 84.6& 97.4& 98.9 & 64.2& 87.1& 92.3 & 524.5\\
    32 & 84.0& 97.3& 98.7 & 64.2& 86.9& 92.4 & 523.0 \\ 

	\bottomrule[1.1pt]
	\end{tabular}
    \caption{Effect  of  the number of MSGAT dynamic attention heads on \textsc{Flickr30K} 1K test set. Best results are in bold.}
	\label{tab:attention heads}
\end{table}

\section{Conclusion}

%In this work, we present MSGAT, an efficient spiking framework for image-text retrieval that combines structure-aware graph reasoning with similarity-space alignment. The proposed Multi-head Spiking Graph Attention Network performs spike-driven adaptive neighborhood aggregation over visual and textual graphs, enabling fine-grained relational modeling under sparse event-driven computation. To address the distribution mismatch between discrete spike representations and continuous global representations, Sim-Fuse integrates coarse- and fine-grained matching relations directly in the similarity space, while the detached fused similarity further guides both branches through KL-divergence alignment. Extensive experiments on Flickr30K and MSCOCO demonstrate that MSGAT consistently outperforms previous SNN retrieval methods and achieves competitive or superior performance compared with state-of-the-art ANN-based approaches. Notably, with only two time steps, MSGAT surpasses its corresponding ANN implementation while reducing theoretical energy consumption by approximately 55\%. Overall, this work advances structure-aware multimodal reasoning in SNNs and provides a promising solution for accurate and energy-efficient cross-modal retrieval.
In this work, we present MSGAT, a spiking framework for image-text retrieval that combines structure-aware graph reasoning with similarity-space alignment. The proposed Multi-head Spiking Graph Attention Network performs spike-driven adaptive neighborhood aggregation over visual and textual graphs, enabling fine-grained relational modeling under sparse event-driven computation. To address the distribution mismatch between discrete spike representations and continuous global representations produced by MSGAT and GPO, respectively, Sim-Fuse integrates coarse- and fine-grained matching relations in similarity space. Experiments on Flickr30K and MSCOCO demonstrate that MSGAT outperforms existing SNN retrieval baseline and surpasses state-of-the-art ANN-based methods. With only two time steps, MSGAT exceeds its ANN counterpart GATv2, while reducing theoretical energy consumption by approximately 55\%. Overall, this work advances structure-aware multimodal reasoning in SNNs and provides a promising, accurate, and energy-efficient solution for cross-modal retrieval.
\bibliography{aaai2027}

@String(AAAI = {AAAI})

@article{third-network-LIF,
  author  = {Wolfgang Maass},
  title   = {Networks of spiking neurons: The third generation of neural network models},
  journal = {Neural Networks},
  volume  = {10},
  number  = {9},
  pages   = {1659--1671},
  year    = {1997}
}

@inproceedings{SCAN,
  author    = {Kuang-Huei Lee and Xi Chen and Gang Hua and Houdong Hu and Xiaodong He},
  title     = {Stacked Cross Attention for Image-Text Matching},
  booktitle = {Proc. Eur. Conf. Comput. Vis.},
  pages     = {212--228},
  year      = {2018}
}

@inproceedings{VSE-infty,
  author    = {Jiacheng Chen and Hexiang Hu and Hao Wu and Yuning Jiang and Changhu Wang},
  title     = {Learning the Best Pooling Strategy for Visual Semantic Embedding},
  booktitle = {Proc. IEEE/CVF Conf. Comput. Vis. Pattern Recog.},
  pages     = {15789--15798},
  year      = {2021}
}

@inproceedings{VSE++,
  author    = {Fartash Faghri and David J. Fleet and Jamie Ryan Kiros and Sanja Fidler},
  title     = {VSE++: Improving Visual-Semantic Embeddings with Hard Negatives},
  booktitle = {Proc. Brit. Mach. Vis. Conf.},
  pages     = {12},
  year      = {2018}
}

@inproceedings{NAAF,
  author    = {Kun Zhang and Zhendong Mao and Quan Wang and Yongdong Zhang},
  title     = {Negative-Aware Attention Framework for Image-Text Matching},
  booktitle = {Proc. IEEE/CVF Conf. Comput. Vis. Pattern Recog.},
  pages     = {15640--15649},
  year      = {2022}
}

@inproceedings{VSRN,
  author    = {Kunpeng Li and Yulun Zhang and Kai Li and Yuanyuan Li and Yun Fu},
  title     = {Visual Semantic Reasoning for Image-Text Matching},
  booktitle = {Proc. IEEE/CVF Int. Conf. Comput. Vis.},
  pages     = {4653--4661},
  year      = {2019}
}

@article{Flickr30K,
  author  = {Peter Young and Alice Lai and Micah Hodosh and Julia Hockenmaier},
  title   = {From image descriptions to visual denotations: New similarity metrics for semantic inference over event descriptions},
  journal = {Trans. Assoc. Comput. Linguistics},
  volume  = {2},
  pages   = {67--78},
  year    = {2014}
}

@inproceedings{MSCOCO,
  author    = {Tsung{-}Yi Lin and Michael Maire and Serge J. Belongie and James Hays and Pietro Perona and Deva Ramanan and Piotr Doll{\'{a}}r and C. Lawrence Zitnick},
  title     = {Microsoft COCO: Common Objects in Context},
  booktitle = {Proc. Eur. Conf. Comput. Vis.},
  pages     = {740--755},
  year      = {2014}
}

@article{Spikingjelly,
author = {Wei Fang and Yanqi Chen and Jianhao Ding and Zhaofei Yu and Timothee Masquelier and Ding Chen and Liwei Huang and Huihui Zhou and Guoqi Li and Yonghong Tian},
title = {SpikingJelly: An open-source machine learning infrastructure platform for spike-based intelligence},
journal = {Science Advances},
volume = {9},
number = {40},
pages = {eadi1480},
year = {2023},

}

@article{Cao_CNN2SNN,
  author  = {Yongqiang Cao and Yang Chen and Deepak Khosla},
  title   = {Spiking Deep Convolutional Neural Networks for Energy-Efficient Object Recognition},
  journal = {Int. J. Comput. Vis.},
  volume  = {113},
  number  = {1},
  pages   = {54--66},
  year    = {2015}
}

@article{wu-STBP,
  author  = {Wu, Yujie and Deng, Lei and Li, Guoqi and Zhu, Jun and Shi, Luping},
  title   = {Spatio-temporal backpropagation for training high-performance spiking neural networks},
  journal = {Front. Neurosci.},
  volume  = {12},
  pages   = {331},
  year    = {2018}
}

@inproceedings{Zhou_spikformer,
  author    = {Zhaokun Zhou and Yuesheng Zhu and Chao He and Yaowei Wang and Shuicheng Yan and Yonghong Tian and Li Yuan},
  title     = {Spikformer: When Spiking Neural Network Meets Transformer},
  booktitle = {Proc. Int. Conf. Learn. Represent.},
  year      = {2023}
}

@inproceedings{QKformer,
  author    = {Chenlin Zhou and Han Zhang and Zhaokun Zhou and Liutao Yu and Liwei Huang and Xiaopeng Fan and Li Yuan and Zhengyu Ma and Huihui Zhou and Yonghong Tian},
  title     = {QKFormer: Hierarchical Spiking Transformer using Q-K Attention},
  booktitle = {Adv. Neural Inform. Process. Syst.},
  year      = {2024}
}

@inproceedings{Spike-Driven-transformer-V2,
  author    = {Man Yao and Jiakui Hu and Tianxiang Hu and Yifan Xu and Zhaokun Zhou and Yonghong Tian and Bo Xu and Guoqi Li},
  title     = {Spike-driven Transformer V2: Meta Spiking Neural Network Architecture Inspiring the Design of Next-generation Neuromorphic Chips},
  booktitle = {Proc. Int. Conf. Learn. Represent.},
  year      = {2024}
}

@inproceedings{LV-SpikeCNN-test,
  author    = {Changze Lv and Jianhan Xu and Xiaoqing Zheng},
  title     = {Spiking Convolutional Neural Networks for Text Classification},
  booktitle = {Proc. Int. Conf. Learn. Represent.},
  year      = {2023}
}

@article{USER,
  author  = {Yan Zhang and Zhong Ji and Di Wang and Yanwei Pang and Xuelong Li},
  title   = {USER: Unified Semantic Enhancement With Momentum Contrast for Image-Text Retrieval},
  journal = {IEEE Trans. Image Process.},
  volume  = {33},
  pages   = {595--609},
  year    = {2024}
}

@inproceedings{CUSA,
  author    = {Hailang Huang and Zhijie Nie and Ziqiao Wang and Ziyu Shang},
  title     = {Cross-Modal and Uni-Modal Soft-Label Alignment for Image-Text Retrieval},
  booktitle = {Proc. AAAI Conf. Artif. Intell.},
  pages     = {18298--18306},
  year      = {2024}
}

@inproceedings{MMCA,
  author    = {Xi Wei and Tianzhu Zhang and Yan Li and Yongdong Zhang and Feng Wu},
  title     = {Multi-Modality Cross Attention Network for Image and Sentence Matching},
  booktitle = {Proc. IEEE/CVF Conf. Comput. Vis. Pattern Recog.},
  pages     = {10938--10947},
  year      = {2020}
}

@article{VSRN++,
  author  = {Kunpeng Li and Yulun Zhang and Kai Li and Yuanyuan Li and Yun Fu},
  title   = {Image-Text Embedding Learning via Visual and Textual Semantic Reasoning},
  journal = {IEEE Trans. Pattern Anal. Mach. Intell.},
  volume  = {45},
  number  = {1},
  pages   = {641--656},
  year    = {2023}
}

@article{FasterRCNN,
  author  = {Shaoqing Ren and Kaiming He and Ross B. Girshick and Jian Sun},
  title   = {Faster R-CNN: Towards Real-Time Object Detection with Region Proposal Networks},
  journal = {IEEE Trans. Pattern Anal. Mach. Intell.},
  volume  = {39},
  number  = {6},
  pages   = {1137--1149},
  year    = {2017}
}

@inproceedings{SeqSNN,
  author    = {Changze Lv and Yansen Wang and Dongqi Han and Xiaoqing Zheng and Xuanjing Huang and Dongsheng Li},
  title     = {Efficient and Effective Time-Series Forecasting with Spiking Neural Networks},
  booktitle = {Proc. Int. Conf. Mach. Learn.},
  year      = {2024}
}

@inproceedings{BERT,
  author    = {Jacob Devlin and Ming-Wei Chang and Kenton Lee and Kristina Toutanova},
  title     = {BERT: Pre-training of Deep Bidirectional Transformers for Language Understanding},
  booktitle = {Proc. Conf. North Am. Chapter Assoc. Comput. Linguist.},
  pages     = {4171--4186},
  year      = {2019}
}

@inproceedings{ONESTEP,
  author    = {Xiaotian Song and Andy Song and Rong Xiao and Yanan Sun},
  title     = {One-step Spiking Transformer with a Linear Complexity},
  booktitle = {Proc. Int. Joint Conf. Artif. Intell.},
  pages     = {3142--3150},
  year      = {2024}
}

@inproceedings{CHAN,
  author    = {Zhengxin Pan and Fangyu Wu and Bailing Zhang},
  title     = {Fine-grained Image-text Matching by Cross-modal Hard Aligning Network},
  booktitle = {Proc. IEEE/CVF Conf. Comput. Vis. Pattern Recog.},
  pages     = {19275--19284},
  year      = {2023}
}

@inproceedings{HREM,
  author    = {Zheren Fu and Zhendong Mao and Yan Song and Yongdong Zhang},
  title     = {Learning Semantic Relationship among Instances for Image-Text Matching},
  booktitle = {Proc. IEEE/CVF Conf. Comput. Vis. Pattern Recog.},
  pages     = {15159--15168},
  year      = {2023}
}

@inproceedings{MaxMatch,
  author    = {Hani Alomari and Anushka Sivakumar and Andrew Zhang and Chris Thomas},
  title     = {Maximal Matching Matters: Preventing Representation Collapse for Robust Cross-Modal Retrieval},
  booktitle = {Proc. Annu. Meet. Assoc. Comput. Linguist.},
  pages     = {31769--31785},
  year      = {2025}
}

@inproceedings{liuwenxuan,
  author    = {Wenxuan Liu and Yao Deng and Kang Chen and Xian Zhong and Zhaofei Yu and Tie-Jun Huang},
  title     = {SOTA: Spike-Navigated Optimal Transport Saliency Region Detection in Composite-bias Videos},
  booktitle = {Proc. Int. Joint Conf. Artif. Intell.},
  year      = {2025}
}

@article{SUYOLO,
  author  = {Chenyang Li and Wenxuan Liu and Guoqiang Gong and Xiaobo Ding and Xian Zhong},
  title   = {SU-YOLO: Spiking Neural Network for Efficient Underwater Object Detection},
  journal = {Neurocomputing},
  volume  = {644},
  pages   = {130310},
  year    = {2025}
}

@inproceedings{spike2former,
  title={Spike2former: Efficient spiking transformer for high-performance image segmentation},
  author={Lei, Zhenxin and Yao, Man and Hu, Jiakui and Luo, Xinhao and Lu, Yanye and Xu, Bo and Li, Guoqi},
  booktitle={Proc. AAAI Conf. Artif. Intell.},
  volume={39},
  number={2},
  pages={1364--1372},
  year={2025}
}

@inproceedings{spikeyolo,
  title={Integer-valued training and spike-driven inference spiking neural network for high-performance and energy-efficient object detection},
  author={Luo, Xinhao and Yao, Man and Chou, Yuhong and Xu, Bo and Li, Guoqi},
  booktitle={Proc. Eur. Conf. Comput. Vis.},
  pages={253--272},
  year={2024},
}

@inproceedings{CoreNLP,
  author       = {Christopher D. Manning and
                  Mihai Surdeanu and
                  John Bauer and
                  Jenny Rose Finkel and
                  Steven Bethard and
                  David McClosky},
  title        = {The Stanford CoreNLP Natural Language Processing Toolkit},
  booktitle    = {Proc. Annu. Meet. Assoc. Comput. Linguist.},
  pages        = {55--60},
  year         = {2014},
  
}

@inproceedings{SpikeGCN,
  author       = {Zulun Zhu and
                  Jiaying Peng and
                  Jintang Li and
                  Liang Chen and
                  Qi Yu and
                  Siqiang Luo},
  title        = {Spiking Graph Convolutional Networks},
  booktitle    = {Proc. Int. Joint Conf. Artif. Intell.},
  pages        = {2434--2440},
  year         = {2022},
  url          = {https://doi.org/10.24963/ijcai.2022/338},
  doi          = {10.24963/IJCAI.2022/338},

}

@inproceedings{SpikeGCL,
  author       = {Jintang Li and
                  Huizhe Zhang and
                  Ruofan Wu and
                  Zulun Zhu and
                  Baokun Wang and
                  Changhua Meng and
                  Zibin Zheng and
                  Liang Chen},
  title        = {A Graph is Worth 1-bit Spikes: When Graph Contrastive Learning Meets
                  Spiking Neural Networks},
  booktitle    = {Proc. Int. Conf. Learn. Represent.},
  publisher    = {OpenReview.net},
  year         = {2024},

}

@article{GSAT,
  author       = {Beibei Wang and
                  Bo Jiang},
  title        = {Spiking GATs: Learning Graph Attentions via Spiking Neural Network},
  journal      = {CoRR},
  volume       = {abs/2209.13539},
  year         = {2022},
  url          = {https://doi.org/10.48550/arXiv.2209.13539},
  doi          = {10.48550/ARXIV.2209.13539},
  eprinttype   = {arXiv},
  eprint       = {2209.13539},

}

@inproceedings{SpikingHAN,
  author       = {Buqing Cao and
                  Qian Peng and
                  Xiang Xie and
                  Liang Chen and
                  Min Shi and
                  Jianxun Liu},
  editor       = {Sven Koenig and
                  Chad Jenkins and
                  Matthew E. Taylor},
  title        = {Spiking Heterogeneous Graph Attention Networks},
  booktitle    = {Proc. AAAI Conf. Artif. Intell.},
  pages        = {19853--19861},
  publisher    = {{AAAI} Press},
  year         = {2026},
  url          = {https://doi.org/10.1609/aaai.v40i24.39068},
  doi          = {10.1609/AAAI.V40I24.39068},
}

@inproceedings{SpikeNet,
  author       = {Jintang Li and
                  Zhouxin Yu and
                  Zulun Zhu and
                  Liang Chen and
                  Qi Yu and
                  Zibin Zheng and
                  Sheng Tian and
                  Ruofan Wu and
                  Changhua Meng},
  title        = {Scaling Up Dynamic Graph Representation Learning via Spiking Neural
                  Networks},
  booktitle    = {Proc. AAAI Conf. Artif. Intell.},
  pages        = {8588--8596},
  publisher    = {{AAAI} Press},
  year         = {2023},
 
}

@inproceedings{CORA,
  author       = {Khoi Pham and
                  Chuong Huynh and
                  Ser{-}Nam Lim and
                  Abhinav Shrivastava},
  title        = {Composing Object Relations and Attributes for Image-Text Matching},
  booktitle    = {Proc. IEEE/CVF Conf. Comput. Vis. Pattern Recog.},
  pages        = {14354--14363},
  publisher    = {{IEEE}},
  year         = {2024},
  url          = {https://doi.org/10.1109/CVPR52733.2024.01361},
  doi          = {10.1109/CVPR52733.2024.01361},
  
}

@article{CMSF,
  title={Brain-Inspired Multimodal Spiking Neural Network for Image-Text Retrieval},
  author={Xintao Zong and Xian Zhong and Wenxuan Liu and Jianhao Ding and Zhaofei Yu and Tiejun Huang},
  journal={ArXiv},
  year={2026},
  volume={abs/2603.26787},
  url={https://api.semanticscholar.org/CorpusID:286961767}
}

@inproceedings{GSMN,
  author       = {Chunxiao Liu and
                  Zhendong Mao and
                  Tianzhu Zhang and
                  Hongtao Xie and
                  Bin Wang and
                  Yongdong Zhang},
  title        = {Graph Structured Network for Image-Text Matching},
  booktitle    = {Proc. IEEE/CVF Conf. Comput. Vis. Pattern Recog.},
  pages        = {10918--10927},
  publisher    = {Computer Vision Foundation / {IEEE}},
  year         = {2020},
 
}

@article{CGMN,
author = {Cheng, Yuhao and Zhu, Xiaoguang and Qian, Jiuchao and Wen, Fei and Liu, Peilin},
title = {Cross-Modal Graph Matching Network for Image-Text Retrieval},
year = {2022},
issue_date = {November 2022},
volume = {18},
number = {4},
issn = {1551-6857},
url = {https://doi.org/10.1145/3499027},
doi = {10.1145/3499027},
journal = {ACM Trans. Multimedia Comput. Commun. Appl.},

}

@inproceedings{hub-loss,
  author       = {Fangyu Liu and
                  Rongtian Ye and
                  Xun Wang and
                  Shuaipeng Li},
  title        = {{HAL:} Improved Text-Image Matching by Mitigating Visual Semantic
                  Hubs},
  booktitle    = {Proc. AAAI Conf. Artif. Intell.},
  pages        = {11563--11571},
  publisher    = {{AAAI} Press},
  year         = {2020},
}

@article{syops,
  title={Training Full Spike Neural Networks via Auxiliary Accumulation Pathway},
  author={Chen, Guangyao and Peng, Peixi and Li, Guoqi and Tian, Yonghong},
  journal={arXiv preprint arXiv:2301.11929},
  year={2023}
}

@inproceedings{GATv2,
  author       = {Shaked Brody and
                  Uri Alon and
                  Eran Yahav},
  title        = {How Attentive are Graph Attention Networks?},
  booktitle    = {Proc. Int. Conf. Learn. Represent.},
  publisher    = {OpenReview.net},
  year         = {2022},
}

@inproceedings{GAT,
  author    = {Veli{\v{c}}kovi{\'c}, Petar and Cucurull, Guillem and Casanova, Arantxa and Romero, Adriana and Li{\`o}, Pietro and Bengio, Yoshua},
  title     = {Graph Attention Networks},
  booktitle = {Proc. Int. Conf. Learn. Represent.},
  year      = {2018}
}

@article{hungarian,
  title={The Hungarian method for the assignment problem},
  author={Kuhn, Harold W.},
  journal={Nav. Res. Logist. Q.},
  volume={2},
  number={1-2},
  pages={83--97},
  year={1955}
}

%\input{ReproducibilityChecklist.tex}

\end{document}